\def\year{2018}\relax
\documentclass[letterpaper]{article} 
\usepackage{aaai18}  
\usepackage{times}  
\usepackage{helvet}  
\usepackage{courier}  
\usepackage{url}  
\usepackage{graphicx}  
\usepackage[boxed,vlined]{algorithm2e}
\usepackage{amsmath,amssymb,amsfonts,dsfont}
\usepackage{booktabs}  
\usepackage{caption}
\usepackage{subcaption}
\usepackage{xspace}
\usepackage{url}

\newcommand{\citeY}[1]{\citeauthor{#1}~\citeyear{#1}}

\newcommand{\insts}[0]{\Pi}
\newcommand{\inst}[0]{\pi}
\newcommand{\instD}[0]{\mathcal{D}}
\newcommand{\pcs}[0]{\mathbf{\Theta}}
\newcommand{\conf}[0]{\mathbf{\theta}}

\newcommand{\hist}[0]{\mathcal{H}}

\newcommand{\run}[0]{i}

\newcommand{\inc}[0]{\conf_\text{inc}}
\newcommand{\incs}[0]{\Theta_\text{inc}}
\newcommand{\incsI}[0]{\Theta^{\insts^i}_{\text{inc}}}
\newcommand{\defc}[0]{\conf_\text{def}}
\newcommand{\challs}[0]{\Theta_\text{chall}}

\newcommand{\algo}[0]{A}

\newcommand{\perf}[0]{\mathbb{R}}
\newcommand{\surro}[0]{\hat{c}}

\newcommand{\smac}{\textit{SMAC}} 
\newcommand{\paramils}{\textit{ParamILS}} 
\newcommand{\gga}{\textit{GGA}} 
\newcommand{\irace}{\textit{irace}}

\newcommand{\react}{\mbox{\textit{ReACT}}}

\newcommand{\spriss}{\textit{SparrowToRiss}\xspace}
\newcommand{\riss}{\textit{Riss}\xspace}
\newcommand{\sparrow}{\textit{Sparrow}\xspace}

\RestyleAlgo{algoruled}
\DontPrintSemicolon
\LinesNumbered
\SetAlgoVlined
\SetFuncSty{textsc}

\SetKwInOut{Input}{Input}
\SetKwInOut{Output}{Output}
\SetKw{Return}{return}

\DeclareMathOperator*{\argmin}{arg\,min}

\newcommand{\citet}[1]{\citeauthor{#1}~(\citeyear{#1})}

\usepackage{tikz}
\usetikzlibrary{shapes.geometric}
\usetikzlibrary{positioning,shapes,shadows,arrows,calc}
\usetikzlibrary{intersections}
\pgfdeclarelayer{background}
\pgfdeclarelayer{foreground}
\pgfsetlayers{background,main,foreground}

\title{Warmstarting of Model-based Algorithm Configuration}
\author{Marius Lindauer \and Frank Hutter\\ 
University of Freiburg\\
$\{$lindauer,fh$\}$@cs.uni-freiburg.de
}

\begin{document}

\maketitle

\begin{abstract}
The performance of many hard combinatorial problem solvers depends strongly on their parameter settings, and since manual parameter tuning is both tedious and suboptimal the AI community has recently developed several algorithm configuration (AC) methods to automatically address this problem.
While all existing AC methods start the configuration process of an algorithm
$\algo$ from scratch for each new type of benchmark instances, here we propose to exploit information about $\algo$'s performance on previous benchmarks in order to warmstart its configuration on new types of benchmarks.
We introduce two complementary ways in which we can exploit this information to warmstart AC methods based on a predictive model.
Experiments for optimizing a flexible modern SAT solver on twelve different instance sets show that our methods often yield substantial speedups over existing AC methods (up to 165-fold) and can also find substantially better configurations given the same compute budget.
\end{abstract}

\section{Introduction}

Many algorithms in the field of artificial intelligence rely crucially on good parameter settings to 
yield strong performance; prominent examples include solvers for many hard
combinatorial problems (e.g., the propositional
satisfiability problem SAT~\cite{hutter-aij17a} or AI
planning~\cite{fawcett-icasp11a})
as well as a wide
range of machine learning algorithms (in particular deep
neural networks~\cite{snoek-nips12a} and automated machine learning
frameworks~\cite{feurer-nips2015a}).
To overcome the tedious and error-prone task of manual parameter tuning for a
given algorithm $A$, algorithm configuration (AC) procedures automatically
determine a parameter configuration of $A$ with low cost (e.g., runtime) on a given
benchmark set.
General algorithm configuration procedures fall into two categories:
model-free approaches, such as \paramils~\cite{hutter-jair09a},
\irace~\cite{lopez-ibanez-orp16} or \gga~\cite{ansotegui-cp09a},
and model-based approaches, such as \smac{}~\cite{hutter-lion11a}
or \gga++~\cite{ansotegui-ijcai15a}.

Even though model-based approaches learn to predict the cost of
different configurations on the benchmark instances at hand, so far all AC
procedures start their configuration process from scratch when presented with
a new set of benchmark instances.
Compared with the way humans exploit information from past benchmark
sets, this is obviously suboptimal. Inspired by the human ability to learn
across different tasks, we propose to use performance measurements for 
an algorithm on previous benchmark sets in order to warmstart its
configuration on a new benchmark set.
As we will show in the experiments,
our new warmstarting methods can substantially speed up AC procedures, by up to a factor
of $165$.
In our experiments, this amounts to spending less than 20 minutes to obtain
comparable performance as could previously be obtained within two days.

\section{Preliminaries}
\label{sec:pre}

\paragraph{Algorithm configuration (AC).}
Formally, given a target algorithm with
configuration space $\pcs$, a probability distribution $\instD$ across problem instances, as well as a cost
metric $c$ to be minimized, the algorithm configuration (AC) problem is to
determine a parameter configuration $\conf^* \in \pcs$ with low expected cost on
instances drawn from $\instD$:
\begin{equation}
\conf^* \in \argmin_{\conf \in \pcs}
\mathbb{E}_{\inst~\sim\instD}[c(\conf,\inst)].
\end{equation}
In practice, $\inst \sim \instD$ is typically approximated by a finite set of
instances $\insts$ drawn from $\instD$.
An example AC problem is to set a SAT solver's parameters to minimize its 
average runtime on a given benchmark set of 
formal verification instances.
Througout the paper, 
we refer to algorithms for solving the AC problem as \emph{AC procedures}.
They execute the target algorithm with different parameter
configurations $\conf \in \pcs$ on different instances $\inst \in \insts$ and
measure the resulting costs $c(\conf, \inst)$. 

\paragraph{Empirical performance models (EPMs).} A core ingredient in
model-based approaches for AC 
is a probabilistic regression model $\surro:
\pcs \times \insts \to \perf$ that is trained based on the cost values $\langle \mathbf{x}= [ \conf, \inst], y = c(\conf, \inst) \rangle$ observed thus far and can be used to
predict the cost of new parameter configurations $\conf \in \pcs$ on new
problem instances $\inst \in \insts$.
Since this regression model predicts empirical algorithm performance (i.e., its cost), 
it is known as an empirical performance model
(EPM; \citeY{leyton-brown-acm09a,hutter-aij14a}). Random forests have been
established as the best-performing type of EPM and are thus used in all current
model-based AC approaches.

For the purposes of this regression model, the instances $\inst$ are
characterized by instance features. 
These features 
reach from simple ones (such as the number of clauses and variables of a SAT formula)
to more complex ones (such as statistics gathered by briefly running a probing
algorithm). Nowadays, informative instance features are available for most hard
combinatorial problems (e.g., SAT~\cite{nudelmann-cp04},
mixed integer programming~\cite{hutter-aij14a},
AI planning~\cite{fawcett-icaps14a}, and answer set
programming~\cite{hoos-tplp14a}).

\begin{algorithm}[tbp]
\Input{Configuration Space $\pcs$, Instances $\insts$, 
Configuration Budget $B$
}
\BlankLine
$\inc,\hist$ $\leftarrow$ initial\_design($\pcs$, $\insts$); \\
\While{$B$ not exhausted} {
  $\surro$ $\leftarrow$ fit EPM based on $\hist$;\\
  $\challs$ $\leftarrow$ select challengers based on $\surro$ and $\hist$;\\ 
  $\inc,\hist$ $\leftarrow$ race($\challs \cup \{\inc\},\insts, \hist$);\\
}
\Return{$\inc$}
\caption{Model-based Algorithm Configuration}
\label{algo:ac}
\end{algorithm}

\paragraph{Model-based algorithm configuration.} 
The core idea of sequential model-based algorithm configuration
is to iteratively fit an EPM based on the cost  
data observed so far and use it to guide the search for well-performing parameter
configurations.
Algorithm~\ref{algo:ac} outlines the model-based algorithm configuration framework, 
similarly as introduced by \citet{hutter-lion11a} for the AC procedure \smac{},
but also encompassing the \gga{}++ approach by \citet{ansotegui-ijcai15a}.
We now discuss this algorithm framework in detail since our warmstarting
extensions will adapt its various elements.

First, in Line 1 a model-based AC procedure runs 
the algorithm to be optimized with configurations
in a so-called \emph{initial design}, keeping track of their costs and
of the best configuration $\inc$ seen so far (the so-called \emph{incumbent}).
It also keeps track of a \emph{runhistory} $\hist$, which contains tuples
$\langle \conf, \inst, c(\conf, \inst) \rangle$ of the cost $c(\conf, \inst)$ obtained
when evaluating configuration $\conf$ on instance $\inst$.
To obtain good anytime performance, by default \smac{} only executes a single
run of a user-defined default configuration $\defc$ on a randomly-chosen
instance as its initial design and uses $\defc$ as its initial incumbent
$\inc$. 
\gga{}++ samples a set of configurations as initial generation
and races them against each other on a subset of the instances.

In Lines~2-5, the AC procedure performs the model-based search.
While a user-specified configuration budget $B$ (e.g., number of algorithm runs
or wall-clock time) is not exhausted, 
it fits a random-forest-based EPM on the existing cost data in $\hist$ (Line~3), 
aggregates the EPM's predictions
over the instances $\insts$ in order to obtain marginal cost predictions 
$\surro(\conf)$ for each configuration $\conf \in \pcs$ and then uses these
predictions in order to select a set of promising configurations $\challs$ to
challenge the incumbent $\inc$ (Line~4) (\smac)
or to generate well-performing offsprings (\gga++).
For this step, a so-called \emph{acquisition function}
trades off exploitation of promising areas of the configuration space versus
exploration of areas for which the model is still uncertain;
common choices are expected improvement~\cite{jones-jgo98a}, upper confidence
bounds~\cite{srninivas-icml10a} or entropy search~\cite{hennig-jmlr12a}.

To determine a new incumbent configuration $\inc$, in Line~5 
the AC procedure races these challengers and the current incumbent by
evaluating them on individual instances $\inst \in \insts$ and adding the
observed data to $\hist$.
Since these evaluations can be computationally costly
the race only evaluates as many instances as
needed per configuration and terminates slow runs early~\cite{hutter-jair09a}.

\section{Warmstarting Approaches for AC}
\label{sec:appr}

\begin{figure}[t]
\centering
\tikzstyle{activity}=[rectangle, draw=black, rounded corners, text centered, text width=6em, fill=white, drop shadow]
\tikzstyle{data}=[rectangle, draw=black, text centered, fill=black!10, text width=8em, drop shadow]
\tikzstyle{myarrow}=[->, thick]
\begin{tikzpicture}[node distance=8em]

	\node (Inst1) [activity, text width=3em] {$\insts^1$};
	\node (Conf1) [activity,right of=Inst1, node distance=6.6em] {Configurator};
	\node (Algo1) [activity,right of=Conf1, text width=5em, node distance=11em] {Algorithm};
	
	\draw[myarrow] (Inst1) -- (Conf1);
	\draw[myarrow] ($(Conf1.east)+(0.0,0.05)$) to [out=30,in=150] node[above] {$\conf, \inst$} ($(Algo1.west)+(0.0,0.05)$);
	\draw[myarrow] ($(Algo1.west)+(0.0,-0.05)$) to [out=-150,in=-30] node[below] {$c(\conf, \inst)$} ($(Conf1.east)+(0.0,-0.05)$);

	\node (Inst2) [activity, text width=3em, below of=Inst1, node distance=7em] {$\insts^2$};
	\node (Conf2) [activity,right of=Inst2, node distance=6.6em] {Configurator};
	\node (Algo2) [activity,right of=Conf2, text width=5em, node distance=11em] {Algorithm};
	
	\draw[myarrow] (Inst2) -- (Conf2);
	\draw[myarrow] ($(Conf2.east)+(0.0,0.05)$) to [out=30,in=150] node[above] {$\conf, \inst$} ($(Algo2.west)+(0.0,0.05)$);
	\draw[myarrow] ($(Algo2.west)+(0.0,-0.05)$) to [out=-150,in=-30] node[below] {$c(\conf, \inst)$} ($(Conf2.east)+(0.0,-0.05)$);

	\node (Inst3) [activity, text width=3em, below of=Inst2, node distance=7em] {$\insts^3$};
	\node (Conf3) [activity,right of=Inst3, node distance=6.6em] {Configurator};
	\node (Algo3) [activity,right of=Conf3, text width=5em, node distance=11em] {Algorithm};
	
	\draw[myarrow] (Inst3) -- (Conf3);
	\draw[myarrow] ($(Conf3.east)+(0.0,0.05)$) to [out=30,in=150] node[above] {$\conf, \inst$} ($(Algo3.west)+(0.0,0.05)$);
	\draw[myarrow] ($(Algo3.west)+(0.0,-0.05)$) to [out=-150,in=-30] node[below] {$c(\conf, \inst)$} ($(Conf3.east)+(0.0,-0.05)$);
	
	\draw[myarrow] ($(Conf1)+(0,-0.8)$) -- (Conf2) node[right, yshift=3.7em] {$\hist^1 :=\langle \conf, \inst, c(\conf,\inst)\rangle$};
	\draw[myarrow] ($(Conf1)+(0,-0.8)$) -- (Conf2) node[left, yshift=3.7em] {$\Theta^{\insts^1}_{\text{inc}}$};
	
	\draw[myarrow] ($(Conf2)+(0,-0.8)$) -- (Conf3) node[right, yshift=3.7em] {$\hist^1 \cup \hist^2$};
	\draw[myarrow] ($(Conf2)+(0,-0.8)$) -- (Conf3) node[left, yshift=3.7em] {$\Theta^{\insts^1}_{\text{inc}} \cup \Theta^{\insts^2}_{\text{inc}}$};

	\begin{pgfonlayer}{background}

    	\path (Inst1 -| Inst1.west)+(-0.1,0.8) node (resUL) {};
    	\path (Algo1.east |- Algo1.south)+(0.1,-0.5) node(resBR) {};
    	\path [rounded corners, dashed, draw=black!50] (resUL) rectangle (resBR);

		\path (Inst2 -| Inst2.west)+(-0.1,0.8) node (resUL2) {};
    	\path (Algo2.east |- Algo2.south)+(0.1,-0.5) node(resBR2) {};
    	\path [rounded corners, dashed, draw=black!50] (resUL2) rectangle (resBR2);
		
		\path (Inst3 -| Inst3.west)+(-0.1,0.8) node (resUL3) {};
    	\path (Algo3.east |- Algo3.south)+(0.1,-0.5) node(resBR3) {};
    	\path [rounded corners, dashed, draw=black!50] (resUL3) rectangle (resBR3);

    \end{pgfonlayer}
	
\end{tikzpicture}
\caption{Control flow of warmstarting information}
\end{figure}
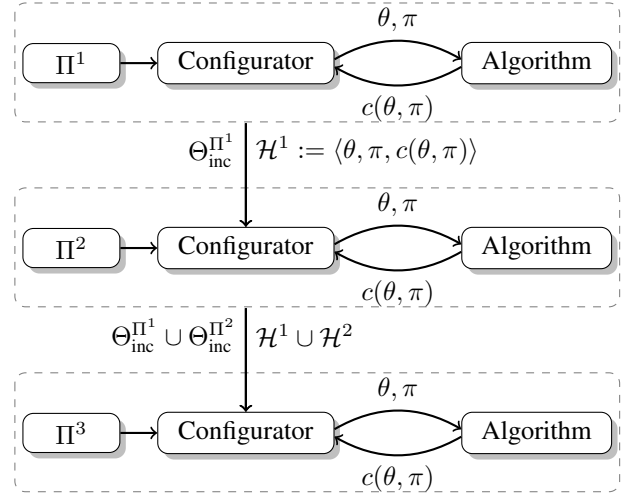

In this section, we discuss how the efficiency of model-based AC procedures
(as described in the previous section) can be improved by warmstarting the
search from data generated in previous AC runs.
We assume that the algorithm to be optimized and its configuration space $\pcs$
is the same in all runs, but the set of instances $\insts$ can change between
the runs.
To warmstart a new AC run, we consider the following data from previous AC runs on previous instance sets $\insts^i$:

\begin{itemize}
  \item Sets of optimized configurations $\incsI$ found in previous AC runs on $\insts^i$---potentially, 
  multiple runs were performed on the same instance set to return the result with best training performance such that $\incsI$ contains the final incumbents from each of these runs;
  \item We denote the union of previous instances as \mbox{$\insts' := \bigcup_{i \in \mathcal{I}} \insts^i$} for set superscripts $i \in \mathcal{I}$.
  \item Runhistory data $\hist' := \bigcup_{i \in \mathcal{I}} \hist^{i}$ of all AC runs on previous instance sets $\insts^i$. \footnote{If the set of instances $\insts$
  and the runhistory $\hist$ are not indexed, we always refer to the ones of the current AC run.}
\end{itemize}

To design warmstarting approaches, we consider the following desired properties:

\begin{enumerate}
  \item When the performance data gathered on previous instance sets is informative about performance on the current instance set, it should speed up our method.
  \item When said performance data is misleading, our method should stop using it and should not be much slower than without it.
  \item The runtime overhead generated by using the prior data should be fairly small.
\end{enumerate}

In the following subsections, we describe different warmstarting
approaches that satisfy these properties.

\subsection{Warmstarting Initial Design (INIT)}

The first approach we consider for warmstarting our model-based AC procedure is
to adapt its initial design (Line~1 of Algorithm~\ref{algo:ac}) to start
from configurations that performed well in the past. Specifically, we include
the incumbent configurations $\incsI$ from all previous AC runs as well as the
user-specified default $\defc$. 

Evaluating all previous incumbents $\incsI$ in the initial design can be
inefficient (contradicting Property~3), 
particularly if they are very similar.
This can happen when the previous instance sets are quite similar, or when multiple 
runs were performed on a single instance set.

To obtain a complementary set of configurations that covers all previously
optimized instances well but is not redundant, we propose to use a two step approach.
First, we determine the best configuration for each previous~$\insts^i$.

\begin{equation}
\incs := \bigcup_{i\in \mathcal{I}} \argmin_{\conf \in \incsI} \sum_{\inst \in \insts^i} c(\conf, \inst)
\end{equation}
   
Secondly,
we use an iterative, greedy forward search to select a complementary set of configurations
across all previous instance sets---inspired by the per-instance selection procedure \emph{Hydra}~\cite{xu-aaai10a}.
Specifically, for the second step 
we define the \emph{mincost} $\tilde{c}(\Theta_j)$ of a set of
configurations $\Theta_j$ on the union of all previous instances $\insts'$ as
\begin{equation}
\tilde{c}(\Theta_j) := \frac{1}{|\insts'|}
\sum_{\inst \in \insts'} \min_{\conf \in \Theta_j} c(\conf,\inst),
\end{equation}
start with $\Theta_1 := \{\defc\}$, and at each iteration, add the configuration
$\conf' \in \incs$ to $\Theta_j$ that minimizes $\tilde{c}(\Theta_j \cup
\{\conf'\})$.
Because $\tilde{c}(\cdot)$ is a
supermodular set function this greedy algorithm is guaranteed to select a set of
configurations whose mincost is within a factor of $(1- 1/e) \approx 0.63 $ of
optimal among sets of the same size~\cite{krause2012submodular}.

Since we do not necessarily know the empirical cost of all $\conf' \in \incs$
on all $\inst \in \insts'$, we use an EPM \mbox{$\surro: \pcs \times \insts
\to \perf$} as a plug-in estimator to predict these costs.
We train this EPM on all previous runhistory data $\hist'$.
In order to enable this, the benchmark sets for all previous AC runs have to be
characterized with the same set of instance features.

In \smac{}, we use this set of complementary configurations
in the initial design using the same racing function as in comparing challengers to the incumbent
(Line~$5$) to obtain the initial incumbent; to avoid rejecting challengers too quickly, a challenger is compared on at least $3$ instances before it can be rejected.
In \gga{}++, these configurations can be included in the first generation of configurations.

\subsection{Data-Driven Model-Warmstarting (DMW)}

Since model-based AC procedures are guided by their EPM,
we considered to warmstart this EPM by including all cost data 
$\hist'$ gathered in previous AC runs as part of its training data.
In the beginning, the predictions of this EPM would mostly rely on $\hist'$, and as more data is acquired on the current benchmark
this would increasingly affect the model.

However, this approach has two disadvantages: 
\begin{enumerate}
  \item When a lot of warmstarting data is available it requires many evaluations on the current instance set to affect model predictions. If the previous data is misleading, this would violate our desired Property~2.   
  \item Fitting the EPM 
  on $\hist \cup \hist'$ will be expensive even in early iterations, because $\hist'$ will typically contain many observations. Even by using \smac{}'s mechanism to invest at least the same amount of time in Lines~3 and~4 as in Line~5, in preliminary experiments this slowed down \smac{} substantially (violating Property~3). 
\end{enumerate}
For these two reasons, we do not use this approach for warmstarting but propose an alternative. 
Specifically, to avoid the computational overhead of refitting a very large EPM in each iteration,
and to allow our model to discard misleading previous data, we propose to fit 
individual EPMs $\surro_{i}$ for each $\hist^i$ once and 
to combine their predictions with those of an EPM $\surro$ fitted on the
newly gathered cost data $\hist$.
This relates to stacking in ensemble learning~\cite{wolpert-nn92a};
however in our case, each constituent EPM is trained on a different dataset. 
Hence, in principle we could even use different instance features
for each instance set.

To aggregate predictions of the individual EPMs, we propose to use a linear
combination:
\begin{equation}
\surro_{\text{DMW}}(\conf,\inst) :=  w_0 + w_{\surro} \cdot \surro(\conf,\inst) + \sum_{i \in \mathcal{I}}{w_\run\cdot \surro_\run(\conf, \inst)}
\end{equation}
\noindent{}where $w$ are weights fitted with stochastic gradient descent (SGD) to
minimize the combined model's root mean squared error (RMSE).
To avoid overfitting of the weights,
we randomly split the current $\hist$ 
into a training and validation set ($2:1$),
use the training set to fit $\surro$,
and then compute predictions of $\surro$ and each $\surro_\run$ on the
validation set, which are used to fit the weights $\mathbf{w}$.
Finally, we re-fit the EPM $\surro$ on all data in $\hist$
to 
obtain a maximally informed model. 

In the beginning of a new AC run, with few data in $\hist$, $\surro$~will not be
very accurate, causing its weight $w_{\surro}$ to be low, such that the previous models
$\surro_i$ will substantially influence the cost predictions.
As more data is gathered in $\hist$, the predictive accuracy of $\surro$ will
improve and the predictions of the previous models $\surro_i$ will become less
important.

Besides weighting based on the accuracy of the individual models,
the weights have the second purpose of scaling the individual model's 
predictions appropriately: these scales reflect the different hardnesses of the
instance sets they were trained on and by setting the weights to minimize RMSE
of the combined model on the current instances $\insts$, they will automatically normalize for scale.

The performance predictions of DMW can be used in any model-based AC procedure, 
such as \smac{} and \gga{}++.

\subsection{Combining INIT and DMW (IDMW)}

Importantly, the two methods we propose are complementary. 
A warmstarted initial design (INIT) can be easily combined with 
data-driven model-warmstarting (DMW) because both approaches 
affect different parts of model-based algorithm configuration: where to start from
and how to integrate the full performance data from the current and the previous benchmarks to decide where to sample next.
In fact, the two warmstarting methods can even synergize to yield more than the sum of their pieces:
by evaluating strong configurations from previous AC runs in the initial design through INIT,
the weights of the stacked model in DMW can be fitted on these important observations early on, improving the accuracy of its
predictions even in early iterations.
  
\section{Experiments}
\label{sec:exp}

We evaluated how our three warmstarting approaches improve the state-of-the-art AC procedure \smac{}.\footnote{The source code of \gga{}++ is not publicly available and thus, we could not run experiments on \gga++.}
In particular, we were interested in the following research questions:
\begin{description}
  \item[Q1] Can warmstarted \smac{} find better performing configurations within the same configuration budget? 
  \item[Q2] Can warmstarted \smac{} find well-performing configurations
  faster than default \smac{}?
  \item[Q3] What is the effect of using warmstarting data $\hist^\run$ from related and unrelated benchmarks?   
\end{description}

\paragraph{Experimental Setup}

To answer these questions, we ran \smac{} (0.5.0) and our warmstarting
variants\footnote{Code and data is publicly available at: \url{http://www.ml4aad.org/smac/}.} 
on twelve well-studied AC tasks from the configurable SAT solver challenge~\cite{hutter-aij17a}, which are publicly available in the algorithm configuration library~\cite{hutter-lion14a}.
Since our warmstarting approaches have to generalize across different instance sets
and not across algorithms,
we considered AC tasks of the highly flexible and robust SAT solver \spriss{} across $12$ instance sets.
\spriss{} is a combination of two well-performing solvers:
\riss~\cite{riss} is a tree-based solver that performs well on industrial and hand-crafted instances;
\sparrow~\cite{sparrow} is a local-search solver that performs well on random, satisfiable instances.
\spriss{} first runs \sparrow{} for a parametrized amount of time
and then runs \riss{} if \sparrow{} could not find a satisfying assignment.
Thus, \spriss can be applied to a large variety of different SAT instances.
\riss, \sparrow and \spriss also won several medals in the international SAT competition.
Furthermore, configuring \spriss{} is a challenging task 
because it has a very large configuration space with $222$ parameters and $176$ conditional dependencies. 

To study warmstarting on different categories of instances,
the AC tasks consider SAT instances from applications with a 
lot of internal structure, hand-crafted instances with some internal structure,
and randomly-generated SAT instances with little structure.
We ran \spriss{} on

\begin{itemize}
  \item application instances from bounded-model checking~(\emph{BMC}), hardware verification (\emph{IBM}) 
and fuzz testing based on circuits (\emph{CF});
  \item hand-crafted instances from graph-isomorphism (\emph{GI}),
low autocorrelation binary sequence (\emph{LABS})
and $n$-rooks instances (\emph{N-Rooks});
  \item randomly generated instances, specifically, 3-SAT instances at
the phase transition from the ToughSAT instance generator
(\emph{3cnf}), a mix of satisfiable and unsatisfiable 3-SAT instances at the phase transition
(\emph{K3}), and unsatisfiable 5-SAT instances from a generator used in the SAT
Challenge $2012$ and SAT Competition $2013$ (\emph{UNSAT-k5}); and on
\item randomly generated satisfiable instances, specifically,
instances with 3 literals per clause and $1000$ clauses (\emph{3SAT1k}),
instances with 5 literals per clause and $500$ clauses (\emph{5SAT500})
and instances with $7$ literals per clause and $90$ clauses (\emph{7SAT90}). 
\end{itemize}   

Further details on
these instances are given in the description of the
configurable SAT solver challenge~\cite{hutter-aij17a}.
The instances were split into a training set for configuration and a test set to
validate the performance of the configured \spriss on unseen instances.

For each configuration run on a benchmark set in one of the categories, 
our warmstarting methods had
access to observations on the other two benchmark sets in the category.
For example, warmstarted \smac{} optimizing \spriss on \emph{IBM} had access to
the observations and final incumbents of \spriss on \emph{CF} and \emph{BMC}.
 
As a cost metric, we chose the commonly-used penalized average runtime metric
(PAR$10$, i.e., counting each timeout as $10$ times the runtime cutoff) with a cutoff of $300$ CPU seconds.
To avoid a constant inflation of the PAR$10$ values,
we removed all test instances post hoc that were never solved by any
configuration in our experiments (%
$11$ \emph{CF} instances, 
$69$ \emph{IBM} instances, 
$17$ \emph{BMC} instances, 
$21$ \emph{GI} instances,
$72$ \emph{LABS} instances and
$73$ \emph{3cnf} instances).


On each AC task, we ran 10 independent \smac{} runs with a configuration
budget of $2$ days each. All runs were run on a compute cluster 
with nodes equipped with two Intel Xeon E5-2630v4 and $128$GB memory
running CentOS 7.

\begin{table*}[t]
\centering
\begin{tabular}{l|rrr|rrr||lrrrr}
\toprule
			& \multicolumn{6}{|c||}{PAR10 scores} & \multicolumn{5}{c}{Speedup over default SMAC}\\
		   &   $\defc$ & SMAC  & AAF   & INIT   & DMW  & IDMW &   & AAF   & INIT               & DMW                           & IDMW     \\
\midrule
 CF        &    326.5 & $\mathbf{125.8}$             & $140.0$          & $\mathbf{126.6}$ & $\mathbf{122.0}$            & $\mathbf{\underline{116.4}}$ & &  0.1 &    0.5 &   0.7 &        \textbf{2.7} \\
 IBM       &    150.6  & $\mathbf{50.6}$              & $49.0$           & $\mathbf{47.8}$  & $\mathbf{\underline{47.5}}$ & $48.8$   & &   3.9 &   \textbf{16.2} &   1.4 &        9                         \\
 BMC       &    421.5 & $209.6$                      & $230.7$          & $203.1$          & $\mathbf{155.9}$            & $\mathbf{\underline{137.4}}$ &  & 1.2 &    1   &  11   &       \textbf{29.3}\\
\midrule
 GI        &    314.1 & $\mathbf{165.0}$             & $\mathbf{165.6}$ & $\mathbf{165.6}$ & $\mathbf{165.4}$            & $\mathbf{\underline{152.7}}$ &   &\textbf{25.6} &    0.6 &   7.1 &       19.4\\
 LABS      &    330.1 & $\mathbf{\underline{232.9}}$ & $291.2$          & $271.2$          & $285.9$                     & $286.7$   &  & 0.8 &    0.8 &   0.8 &        0.8                    \\
 N-Rooks   &    116.7 & $\mathbf{\underline{8.6}}$   & $18.1$           & $27.3$           & $27.5$                      & $\mathbf{12.7}$       &    &  0.4 &    0.4 &   0.4 &        0.5        \\
\midrule
 3cnf &    890.5 & $890.5$                      & $\mathbf{822.8}$ & $\mathbf{877.5}$ & $890.3$                     & $\mathbf{\underline{812.8}}$ & &  \textbf{10.7} &    1   &   1   &        8.4\\
 K3        &    152.8 & $\mathbf{30.1}$              & $53.9$           & $\mathbf{42.9}$  & $\mathbf{39.9}$             & $\mathbf{\underline{29.7}}$ &  &  0.9 &    0.9 &   1.8 &        \textbf{1.8}\\
 UNSAT-k5  &    151.9 & $\mathbf{\underline{1.1}}$   & $1.2$            & $\mathbf{1.1}$   & $1.3$                       & $1.2$             &  & 1   &    1   &   1   &        1               \\
\midrule
 3SAT1k    &    104.4 & $\mathbf{76.6}$              & $\mathbf{75.2}$  & $\mathbf{75.2}$  & $\mathbf{82.6}$             & $\mathbf{\underline{69.8}}$ & &  3.1 &    2.1 &   2.1 &        \textbf{3.8}\\
 5SAT500   &   3000    & $20.5$                       & $14.6$           & $\mathbf{14.7}$  & $\mathbf{\underline{8.3}}$  & $\mathbf{9.6}$        &    &  \textbf{6}   &    0.7 &   0.7 &        0.8    \\
 7SAT90    &     52.3 & $38.0$                       & $31.7$           & $\mathbf{20.2}$  & $\mathbf{\underline{19.7}}$ & $\mathbf{31.2}$       &   &  53.5 &    2.3 &   0.5 &      \textbf{165.3}     \\
 \midrule
 \multicolumn{7}{c||}{}&$\varnothing$  &       2.4 &    1.1 &   1.3 &        4.3\\
\bottomrule
\end{tabular}
\caption{\textbf{Left:} PAR10 score (sec) of $\defc$, i.e., the default configuation of \spriss, and the final \spriss configurations returned by the different \smac{} variants; median across $10$ \smac{} runs. The ``SMAC'' column shows the performance of default \smac{} without warmstarting. 
 Best PAR10 is underlined and we highlighted runs in bold face for which there is no statistical evidence according to a
 (one-sided) Mann-Whitney U test ($\alpha=0.05$) that they performed worse than the best configurator.
 \textbf{Right:}
 Speedup of warmstarted \smac{} compared to default \smac{}.
 This is computed by comparing the time points of \smac{} with and without warmstarting
after which they do not perform significantly worse
(according to a permutation test) than \smac{} with the full budget.
Speedups $>1$ indicate that warmstarted \smac{} reached the final performance of default \smac{} faster, speedups
$<1$ indicate that default \smac{} was faster.
We marked the best speedup ($>1$) in bold-face.
The last row shows the geometric average across all speedups.
\label{tab:perf}
}
\end{table*}

\subsection{Baselines}
As baselines, we ran (I) the user-specified default configuration $\defc$ to show the effect of algorithm configuration,
(II) \smac{} without warmstarting,
and (III) a state-of-the-art warmstarting approach for hyperparameter optimizers proposed by \citet{wistuba-dsaa16}, which we abbreviate as ``adapted acquisition function'' (AAF). 
The goal of AAF is to bias the acquisition function (Line~$4$ in Algorithm~\ref{algo:ac}) towards previously well-performing regions in the configuration space.\footnote{We note that combining AAF and INIT is not effective because evaluating the incumbents of INIT would nullify the acquisition function bias of AAF.} 
To generalize AAF to algorithm configuration, we use marginalized prediction across all instances $\hat{c}(\conf) := \frac{1}{|\insts|}\sum_{\inst \in \insts} \hat{c}(\conf, \inst)$.

\subsection{Q1: Same configuration Budget}

The left part of Table~\ref{tab:perf} shows the median PAR10 test scores of the finally-returned configurations $\inc$
across the $10$ \smac{} runs.
Default \smac{} nearly always improved the PAR$10$ scores of \spriss substantially
compared to the \spriss default, yielding up to a $138$-fold speedup (on \emph{UNSAT-k5}).
Warmstarted \smac{} performed significantly better yet on $4$ of the AC tasks
(\emph{BMC}, \emph{3cnf}, \emph{5SAT500} and \emph{7SAT90}), with additional speedups up to
$2.1$-fold (on \emph{5SAT500}).
On two of the crafted instance sets (\emph{LABS} and \emph{N-Rooks}),
the warmstarting approaches performed worse than default \smac{}---details discussed later.

Overall, the best results were achieved by the combination of our approaches, IDMW.
This yielded the best performance of all approaches in 6 of the 12 scenarios (with sometimes substantial improvements over default SMAC) and
statistically insignificantly different results than the best approach in 3 of the scenarios.
Notably, IDMW performed better on average than its individual components INIT and DMW
and clearly outperformed AAF.  

\subsection{Q2: Speedup}

\begin{figure*}[tbh]
\centering
\begin{subfigure}[b]{0.325\textwidth}
	\includegraphics[width=\textwidth]{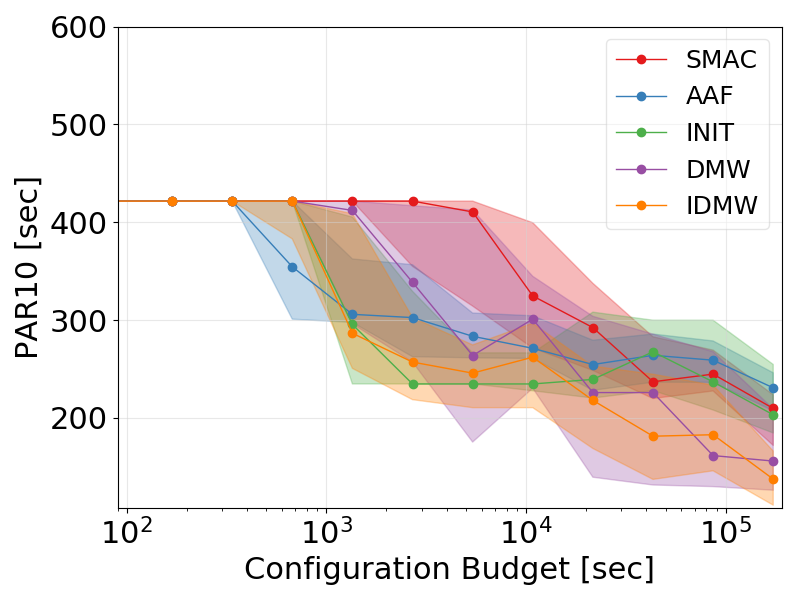}
	\caption{BMC}
	\label{fig:sp_bmc}
\end{subfigure}
\begin{subfigure}[b]{0.325\textwidth}
	\includegraphics[width=\textwidth]{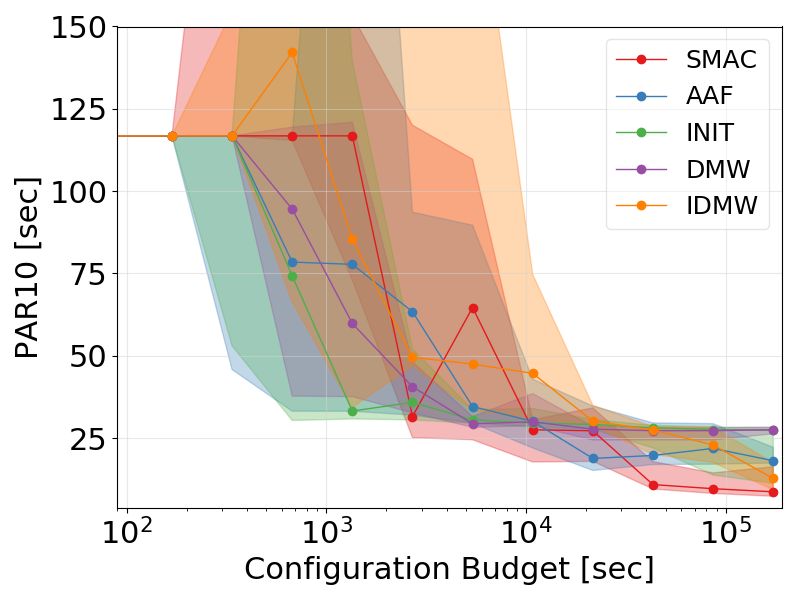}
	\caption{N-Rooks}
	\label{fig:sp_queens}
\end{subfigure}
\begin{subfigure}[b]{0.325\textwidth}
	\includegraphics[width=\textwidth]{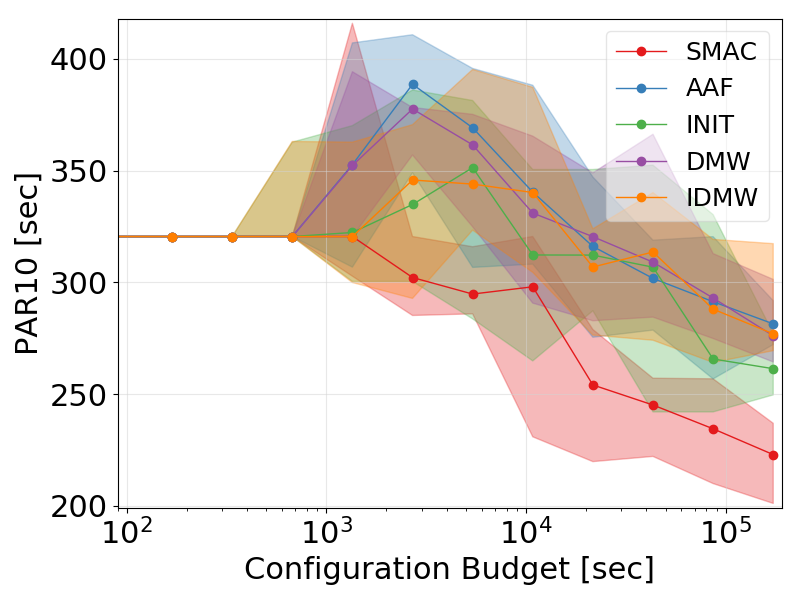}
	\caption{LABS}
	\label{fig:sp_labs}
\end{subfigure}
\caption{Median PAR10 of \spriss over configuration time with $25\%$ and $75\%$ percentiles as uncertainties.}
\label{fig:over_time}
\end{figure*}

The right part of Table~\ref{tab:perf} shows how much faster our warmstarted
\smac{} reached the PAR10 performance default that \smac{} reached with
the full configuration budget.\footnote{A priori it is not clear how to define a speedup metric comparing algorithm configurators across several runs.
To take noise into account across our $10$ runs,
we performed a permutation test (with $\alpha=0.05$ with $10\ 000$ permutations) to determine the first
time point from which onwards there was no statistical evidence that
default \smac{} with a full budget would perform better.
To take early convergence/stagnation of default \smac{} into account,
we compute the speedup of default \smac{} to itself
and divide the speedups by default \smac{}'s speedup.}
The warmstarting methods outperformed default SMAC in almost all cases (again except \emph{LABS} and \emph{N-Rooks}), with up to 165-fold speedups. 
The most consistent speedups were achieved by the combination of our warmstarting approaches, IDMW, with a geometric-average $4.3$-fold speedup.
We note that our baseline AAF also yielded good speedups (geometric average of 2.4), but its final performance was often quite poor (see left part of Table~\ref{tab:perf}).

Figure~\ref{fig:over_time} illustrates the anytime test performance of all \smac{} variants.\footnote{Since Figure~\ref{fig:over_time} shows test performance on unseen test instances, performance is not guaranteed to improve monotonically (a new best configuration on the training instances might not generalize well to the test instances).}
In Figure~\ref{fig:sp_bmc}, AAF, INIT and IDMW improved the performance of \spriss very early (after roughly $700$-$1000$ seconds), but only the DMW variants performed well in the long run.

To study the effect of our worst results,
Figure~\ref{fig:sp_queens} and \ref{fig:sp_labs} show the anytime performance on 
\emph{N-Rooks} and \emph{LABS}, respectively.
Figure~\ref{fig:sp_queens} shows that warmstarted \smac{} performed better in the beginning,
but that default \smac{} performed slightly better in the end.  
The better initial performance is not captured in our quantitative analysis in Table~\ref{tab:perf}.
In contrast, Figure~\ref{fig:sp_labs} shows that for \emph{LABS}, warmstarted \smac{} was initially mislead and then started improving like default SMAC, but with a time lag; we note that we only observed this pattern on \emph{LABS} and conclude that 
configurations found on \emph{N-Rooks} and \emph{GI} do not generalize to \emph{LABS}.

\subsection{Q3: Warmstarting Influence}

\begin{figure}[tb]
\centering
\begin{subfigure}[b]{0.23\textwidth}
	\includegraphics[width=\textwidth]{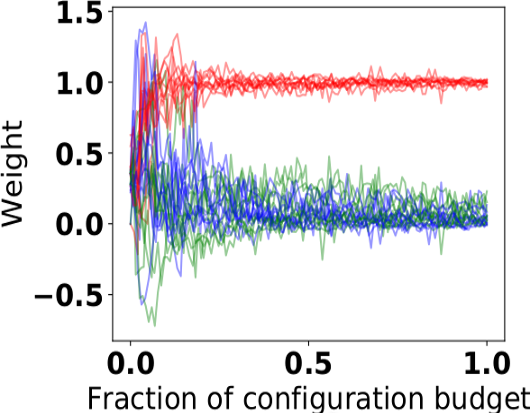}
	\caption{red: IBM, blue: CF,\\ green: BMC}
	\label{fig:weights:ibm}%
\end{subfigure}
\begin{subfigure}[b]{0.23\textwidth}
	\includegraphics[width=\textwidth]{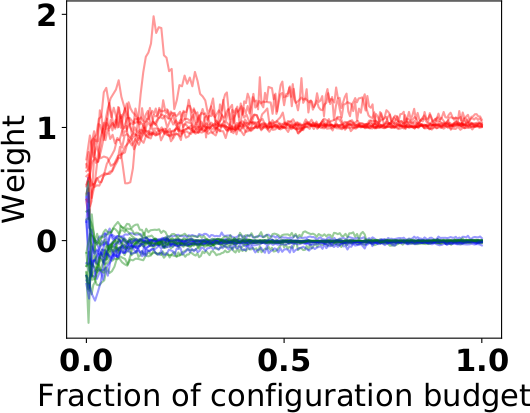}
	\caption{red: UNSAT-k5, blue: K3, green: 3cnf}
	\label{fig:weights:unsat} 
\end{subfigure}
\caption{Weights over time of all $10$ runs \smac{}+DMW. The red curve is the weight on EPM $\surro$ on the current instances; the blue and green curves corresponds to weights on EPMs based on previously optimized instances.}
\label{fig:weights}
\end{figure}

To study how our warmstarting methods learn from previous data,
in Figure~\ref{fig:weights} we show how the weights of the DMW approach changed over time.
Figure~\ref{fig:weights:ibm} shows a representative plot:
the weights were similar in the beginning (i.e., all EPMs contributed
similarly to cost predictions) and over time, the weights of the previous models
decreased, with the weight of the current EPM dominating.
When optimizing on \emph{IBM}, the EPM trained on observations from \emph{CF}
was the most important EPM in the beginning. 

In contrast, Figure~\ref{fig:weights:unsat} shows a case in which the previous performance data acquired for benchmarks \emph{K3} and \emph{3cnf} do not help for cost predictions on \emph{UNSAT-k5}. (This was to be expected, because \emph{3cnf} comprises only satisfiable instances,  \emph{K3} a mix of satisfiable and unsatisfiable instances, and \emph{UNSAT-k5} only unsatisfiable instances.)
As the figure shows, our DMW approach briefly used the data from the mixed \emph{K3} benchmark (blue curves), but quickly focused only on data from the current benchmark. 
These two examples illustrate that our DMW approach indeed successfully used data from related benchmarks and quickly ignored data from unrelated ones.   

\section{Related Work}
\label{sec:rel}

The most related work comes from the field of hyperparameter optimization (HPO)
of machine learning algorithms. HPO, when cast as the optimization of
(cross-)validation error, is a special case of AC. This special case does
not require the concept of problem instances,
does not require the modelling of runtimes of
randomized algorithms, does not need to adaptively terminate slow algorithm
runs and handle the resulting censored algorithm runtimes, and typically deals
with fairly low-dimensional and all-continuous \mbox{(hyper-)parameter}
configuration spaces. 
These works therefore do not directly transfer to the general AC problem.

Several warmstarting approaches exist for HPO. A prominent approach is to learn
surrogate models across
datasets~\cite{swersky-nips13,bardenet-icml13a,yogatama-aistats2014}. All of
these works are based on Gaussian process models whose computational complexity
scales cubically in the number of data points, and therefore, all of them were
limited to hundreds or at most thousands of data points.
We generalize them to the AC setting (which, on top of the differences to
HPO stated above, also needs to handle up to a million cost
measurements for an algorithm) in our DMW approach.

Another approach for warmstarting HPO is by adapting the initial
design.
\citet{feurer-aaai15a} proposed to
initialize HPO in the automatic machine learning framework
\textit{Auto-Sklearn} with well-performing configurations from previous datasets.
They had optimized configurations from $57$ different machine learning data sets
available as warmstarting data and chose which of these to use for a new
dataset based on its characteristics; specifically, they used the optimized
configurations from the $k$ most similar datasets. This approach could be
adapted to AC warmstarting in cases where we have many AC benchmarks.
However, one disadvantage of the approach is that -- unlike our INIT approach -- it does
not aim for complementarity in the selected configurations.
\citet{wistuba-dsaa15} proposed another approach for
warmstarting the initial design which does not depend on instance features
and is not limited to configurations returned in previous optimization experiments.
They combined surrogate predictions from previous runs
and used gradient descent to determine promising configurations. This
approach is limited to continuous \mbox{(hyper-)parameters} and thus does not
apply to the general AC setting.

One related variant of algorithm configuration is the problem of configuring on
a stream of problem instances that changes over time. The \react{}
approach~\cite{fitzgerald-socs14a} targets this problem setting, keeping track
of configurations that worked well on previous instances. If the
characteristics of the instances change over time, it also adapts the current
configuration by combining observations on previous instances and on new
instances.
In contrast to our setting,
\react{} does not return a single configuration for an instance set
and requires parallel compute resources to run a parallel portfolio all the time.

\section{Discussion \& Conclusion}\label{sec:conclusion}

In this paper, we introduced several methods to warmstart model-based algorithm
configuration (AC) using observations from previous AC experiments on different
benchmark instance sets.
As we showed in our experiments, warmstarting can speed up the configuration
process up to 165-fold and can also improve the configurations finally returned.

While we focused on the state-of-the-art configurator \smac{} in our experiments,
our methods are also applicable to other model-based configurators, such as \gga{}++,
and our warmstarted initial design approach is even applicable to model-free configurators,
such as \paramils{} and \irace{}.
We expect that our results would similarly generalize to these.

A practical limitation of our DMW approach (and thus also of IDMW) is that
the memory consumption grows substantially with each additional EPM 
(at least when using random forests fitted on hundreds of thousands of observations).
We also tried to study warmstarting \smac{} for optimizing \spriss{} on all instance sets
except the one at hand, but unfortunately, the memory consumption exceeded 12GB RAM.
Therefore, one possible approach would be to reduce memory consumption and to use instance features
to select a subset of EPMs constructed on similar instances.

Another direction for future work is to combine warmstarting with parameter
importance analysis~\cite{hutter-icml14a,biedenkapp-aaai17a}, e.g., for
determining important parameters on previous instance sets and focusing the search
on these parameters for a new instance set.
Finally, a promising future direction is to integrate warmstarting 
into iterative configuration procedures, such as   
\textit{Hydra}~\cite{xu-aaai10a}, \textit{ParHydra}~\cite{lindauer-aij16a}, or
\textit{Cedalion}~\cite{seipp-aaai15a}, which construct portfolios
of complementary configurations in an iterative fashion using multiple
AC runs.

\section{Acknowledgements}

The authors acknowledge funding by the DFG (German Research Foundation) under
Emmy Noether grant HU 1900/2-1 and support by the state of Baden-W\"urttemberg through bwHPC
and the DFG through grant no INST~39/963-1 FUGG.

\footnotesize
\bibliographystyle{aaai}
\bibliography{strings,lib,local,proc}

\begin{thebibliography}{}

\bibitem[\protect\citeauthoryear{Ans{\'o}tegui \bgroup et al\mbox.\egroup
  }{2015}]{ansotegui-ijcai15a}
Ans{\'o}tegui, C.; Malitsky, Y.; Sellmann, M.; and Tierney, K.
\newblock 2015.
\newblock Model-based genetic algorithms for algorithm configuration.
\newblock In Yang, Q., and Wooldridge, M., eds., {\em Proceedings of the 25th
  International Joint Conference on Artificial Intelligence (IJCAI'15)},
  733--739.

\bibitem[\protect\citeauthoryear{Ans{\'o}tegui, Sellmann, and
  Tierney}{2009}]{ansotegui-cp09a}
Ans{\'o}tegui, C.; Sellmann, M.; and Tierney, K.
\newblock 2009.
\newblock A gender-based genetic algorithm for the automatic configuration of
  algorithms.
\newblock In Gent, I., ed., {\em Proceedings of the Fifteenth International
  Conference on Principles and Practice of Constraint Programming (CP'09)},
  volume 5732 of {\em Lecture Notes in Computer Science},  142--157.
\newblock Springer-Verlag.

\bibitem[\protect\citeauthoryear{Balint \bgroup et al\mbox.\egroup
  }{2011}]{sparrow}
Balint, A.; Frohlich, A.; Tompkins, D.; and Hoos, H.
\newblock 2011.
\newblock Sparrow2011.
\newblock In {\em Proceedings of {SAT} Competition 2011}.

\bibitem[\protect\citeauthoryear{Bardenet \bgroup et al\mbox.\egroup
  }{2014}]{bardenet-icml13a}
Bardenet, R.; Brendel, M.; K{\'e}gl, B.; and Sebag, M.
\newblock 2014.
\newblock Collaborative hyperparameter tuning.
\newblock In Dasgupta, S., and McAllester, D., eds., {\em Proceedings of the
  30th International Conference on Machine Learning (ICML'13)},  199--207.
\newblock Omnipress.

\bibitem[\protect\citeauthoryear{Biedenkapp \bgroup et al\mbox.\egroup
  }{2017}]{biedenkapp-aaai17a}
Biedenkapp, A.; Lindauer, M.; Eggensperger, K.; Fawcett, C.; Hoos, H.; and
  Hutter, F.
\newblock 2017.
\newblock Efficient parameter importance analysis via ablation with surrogates.
\newblock In {\em Proceedings of the Thirty-First Conference on Artificial
  Intelligence (AAAI'17)},  773--779.

\bibitem[\protect\citeauthoryear{Bonet and Koenig}{2015}]{aaai15}
Bonet, B., and Koenig, S., eds.
\newblock 2015.
\newblock {\em Proceedings of the Twenty-nineth Conference on Artificial
  Intelligence (AAAI'15)}. AAAI Press.

\bibitem[\protect\citeauthoryear{Fawcett \bgroup et al\mbox.\egroup
  }{2011}]{fawcett-icasp11a}
Fawcett, C.; Helmert, M.; Hoos, H.; Karpas, E.; Roger, G.; and Seipp, J.
\newblock 2011.
\newblock Fd-autotune: Domain-specific configuration using fast-downward.
\newblock In Helmert, M., and Edelkamp, S., eds., {\em Working notes of the
  Twenty-first International Conference on Automated Planning and Scheduling
  (ICAPS-11), Workshop on Planning and Learning.}

\bibitem[\protect\citeauthoryear{Fawcett \bgroup et al\mbox.\egroup
  }{2014}]{fawcett-icaps14a}
Fawcett, C.; Vallati, M.; Hutter, F.; Hoffmann, J.; Hoos, H.; and
  Leyton{-}Brown, K.
\newblock 2014.
\newblock Improved features for runtime prediction of domain-independent
  planners.
\newblock In Chien, S.; Minh, D.; Fern, A.; and Ruml, W., eds., {\em
  Proceedings of the Twenty-Fourth International Conference on Automated
  Planning and Scheduling (ICAPS-14)}.
\newblock {AAAI}.

\bibitem[\protect\citeauthoryear{Feurer \bgroup et al\mbox.\egroup
  }{2015}]{feurer-nips2015a}
Feurer, M.; Klein, A.; Eggensperger, K.; Springenberg, J.~T.; Blum, M.; and
  Hutter, F.
\newblock 2015.
\newblock Efficient and robust automated machine learning.
\newblock In Cortes, C.; Lawrence, N.; Lee, D.; Sugiyama, M.; and Garnett, R.,
  eds., {\em Proceedings of the 29th International Conference on Advances in
  Neural Information Processing Systems (NIPS'15)}.

\bibitem[\protect\citeauthoryear{Feurer, Springenberg, and
  Hutter}{2015}]{feurer-aaai15a}
Feurer, M.; Springenberg, T.; and Hutter, F.
\newblock 2015.
\newblock Initializing {B}ayesian hyperparameter optimization via
  meta-learning.
\newblock In Bonet and Koenig \shortcite{aaai15},  1128--1135.

\bibitem[\protect\citeauthoryear{Fitzgerald \bgroup et al\mbox.\egroup
  }{2014}]{fitzgerald-socs14a}
Fitzgerald, T.; O'Sullivan, B.; Malitsky, Y.; and Tierney, K.
\newblock 2014.
\newblock React: Real-time algorithm configuration through tournaments.
\newblock In Edelkamp, S., and Bart{\'{a}}k, R., eds., {\em Proceedings of the
  Seventh Annual Symposium on Combinatorial Search (SOCS'14)}.
\newblock {AAAI} Press.

\bibitem[\protect\citeauthoryear{Hennig and Schuler}{2012}]{hennig-jmlr12a}
Hennig, P., and Schuler, C.
\newblock 2012.
\newblock Entropy search for information-efficient global optimization.
\newblock {\em Journal of Machine Learning Research} 98888(1):1809--1837.

\bibitem[\protect\citeauthoryear{Hoos, Lindauer, and
  Schaub}{2014}]{hoos-tplp14a}
Hoos, H.; Lindauer, M.; and Schaub, T.
\newblock 2014.
\newblock claspfolio 2: Advances in algorithm selection for answer set
  programming.
\newblock {\em Theory and Practice of Logic Programming} 14:569--585.

\bibitem[\protect\citeauthoryear{Hutter \bgroup et al\mbox.\egroup
  }{2009}]{hutter-jair09a}
Hutter, F.; Hoos, H.; Leyton-Brown, K.; and St{\"u}tzle, T.
\newblock 2009.
\newblock Param{ILS}: An automatic algorithm configuration framework.
\newblock {\em Journal of Artificial Intelligence Research} 36:267--306.

\bibitem[\protect\citeauthoryear{Hutter \bgroup et al\mbox.\egroup
  }{2014a}]{hutter-lion14a}
Hutter, F.; L\'{o}pez-Ib\'{a}nez, M.; Fawcett, C.; Lindauer, M.; Hoos, H.;
  Leyton-Brown, K.; and St\"utzle, T.
\newblock 2014a.
\newblock Aclib: a benchmark library for algorithm configuration.
\newblock In Pardalos, P., and Resende, M., eds., {\em Proceedings of the
  Eighth International Conference on Learning and Intelligent Optimization
  (LION'14)}, Lecture Notes in Computer Science,  36--40.
\newblock Springer-Verlag.

\bibitem[\protect\citeauthoryear{Hutter \bgroup et al\mbox.\egroup
  }{2014b}]{hutter-aij14a}
Hutter, F.; Xu, L.; Hoos, H.; and Leyton-Brown, K.
\newblock 2014b.
\newblock Algorithm runtime prediction: Methods and evaluation.
\newblock {\em Artificial Intelligence} 206:79--111.

\bibitem[\protect\citeauthoryear{Hutter \bgroup et al\mbox.\egroup
  }{2017}]{hutter-aij17a}
Hutter, F.; Lindauer, M.; Balint, A.; Bayless, S.; Hoos, H.; and Leyton-Brown,
  K.
\newblock 2017.
\newblock The configurable {SAT} solver challenge ({CSSC}).
\newblock {\em Artificial Intelligence Journal (AIJ)} 243:1--25.

\bibitem[\protect\citeauthoryear{Hutter, Hoos, and
  Leyton-Brown}{2011}]{hutter-lion11a}
Hutter, F.; Hoos, H.; and Leyton-Brown, K.
\newblock 2011.
\newblock Sequential model-based optimization for general algorithm
  configuration.
\newblock In Coello, C., ed., {\em Proceedings of the Fifth International
  Conference on Learning and Intelligent Optimization (LION'11)}, volume 6683
  of {\em Lecture Notes in Computer Science},  507--523.
\newblock Springer-Verlag.

\bibitem[\protect\citeauthoryear{Hutter, Hoos, and
  Leyton-Brown}{2014}]{hutter-icml14a}
Hutter, F.; Hoos, H.; and Leyton-Brown, K.
\newblock 2014.
\newblock An efficient approach for assessing hyperparameter importance.
\newblock In Xing, E., and Jebara, T., eds., {\em Proceedings of the 31th
  International Conference on Machine Learning, (ICML'14)},  754--762.
\newblock Omnipress.

\bibitem[\protect\citeauthoryear{Jones, Schonlau, and
  Welch}{1998}]{jones-jgo98a}
Jones, D.; Schonlau, M.; and Welch, W.
\newblock 1998.
\newblock Efficient global optimization of expensive black box functions.
\newblock {\em Journal of Global Optimization} 13:455--492.

\bibitem[\protect\citeauthoryear{Krause and
  Golovin}{2012}]{krause2012submodular}
Krause, A., and Golovin, D.
\newblock 2012.
\newblock Submodular function maximization.
\newblock {\em Tractability: Practical Approaches to Hard Problems} 3(19):8.

\bibitem[\protect\citeauthoryear{Leyton-Brown, Nudelman, and
  Shoham}{2009}]{leyton-brown-acm09a}
Leyton-Brown, K.; Nudelman, E.; and Shoham, Y.
\newblock 2009.
\newblock Empirical hardness models: Methodology and a case study on
  combinatorial auctions.
\newblock {\em Journal of the ACM} 56(4).

\bibitem[\protect\citeauthoryear{Lindauer \bgroup et al\mbox.\egroup
  }{2017}]{lindauer-aij16a}
Lindauer, M.; Hoos, H.; Leyton-Brown, K.; and Schaub, T.
\newblock 2017.
\newblock Automatic construction of parallel portfolios via algorithm
  configuration.
\newblock {\em Artificial Intelligence} 244:272--290.

\bibitem[\protect\citeauthoryear{L{\'{o}}pez{-}Ib{\'{a}}{\~{n}}ez \bgroup et
  al\mbox.\egroup }{2016}]{lopez-ibanez-orp16}
L{\'{o}}pez{-}Ib{\'{a}}{\~{n}}ez, M.; Dubois-Lacoste, J.; Caceres, L.~P.;
  Birattari, M.; and St{\"{u}}tzle, T.
\newblock 2016.
\newblock The irace package: Iterated racing for automatic algorithm
  configuration.
\newblock {\em Operations Research Perspectives} 3:43--58.

\bibitem[\protect\citeauthoryear{Manthey}{2014}]{riss}
Manthey, N.
\newblock 2014.
\newblock Riss 4.27.
\newblock In Belov, A.; Diepold, D.; Heule, M.; and J{\"a}rvisalo, M., eds.,
  {\em Proceedings of {SAT} Competition 2014: Solver and Benchmark
  Descriptions}, volume B-2014-2 of {\em Department of Computer Science Series
  of Publications B},  65--67.
\newblock University of Helsinki.

\bibitem[\protect\citeauthoryear{Nudelman \bgroup et al\mbox.\egroup
  }{2004}]{nudelmann-cp04}
Nudelman, E.; Leyton-Brown, K.; Devkar, A.; Shoham, Y.; and Hoos, H.
\newblock 2004.
\newblock Understanding random {SAT}: Beyond the clauses-to-variables ratio.
\newblock In Wallace, M., ed., {\em Proceedings of the 10th International
  Conference on Principles and Practice of Constraint Programming (CP'04)},
  volume 3258 of {\em Lecture Notes in Computer Science},  438--452.
\newblock Springer-Verlag.

\bibitem[\protect\citeauthoryear{Seipp \bgroup et al\mbox.\egroup
  }{2015}]{seipp-aaai15a}
Seipp, J.; Sievers, S.; Helmert, M.; and Hutter, F.
\newblock 2015.
\newblock Automatic configuration of sequential planning portfolios.
\newblock In Bonet and Koenig \shortcite{aaai15},  3364--3370.

\bibitem[\protect\citeauthoryear{Snoek, Larochelle, and
  Adams}{2012}]{snoek-nips12a}
Snoek, J.; Larochelle, H.; and Adams, R.~P.
\newblock 2012.
\newblock Practical {B}ayesian optimization of machine learning algorithms.
\newblock In Bartlett, P.; Pereira, F.; Burges, C.; Bottou, L.; and Weinberger,
  K., eds., {\em Proceedings of the 26th International Conference on Advances
  in Neural Information Processing Systems (NIPS'12)},  2960--2968.

\bibitem[\protect\citeauthoryear{Srinivas \bgroup et al\mbox.\egroup
  }{2010}]{srninivas-icml10a}
Srinivas, N.; Krause, A.; Kakade, S.; and Seeger, M.
\newblock 2010.
\newblock {G}aussian process optimization in the bandit setting: No regret and
  experimental design.
\newblock In F{\"u}rnkranz, J., and Joachims, T., eds., {\em Proceedings of the
  27th International Conference on Machine Learning (ICML'10)},  1015--1022.
\newblock Omnipress.

\bibitem[\protect\citeauthoryear{Swersky, Snoek, and
  Adams}{2013}]{swersky-nips13}
Swersky, K.; Snoek, J.; and Adams, R.
\newblock 2013.
\newblock Multi-task {B}ayesian optimization.
\newblock In Burges, C.; Bottou, L.; Welling, M.; Ghahramani, Z.; and
  Weinberger, K., eds., {\em Proceedings of the 27th International Conference
  on Advances in Neural Information Processing Systems (NIPS'13)},  2004--2012.

\bibitem[\protect\citeauthoryear{Wistuba, Schilling, and
  Schmidt-Thieme}{2015}]{wistuba-dsaa15}
Wistuba, M.; Schilling, N.; and Schmidt-Thieme, L.
\newblock 2015.
\newblock Learning hyperparameter optimization initializations.
\newblock In {\em Proceedings of the International Conference on Data Science
  and Advanced Analytics (DSAA)},  1--10.
\newblock {IEEE}.

\bibitem[\protect\citeauthoryear{Wistuba, Schilling, and
  Schmidt{-}Thieme}{2016}]{wistuba-dsaa16}
Wistuba, M.; Schilling, N.; and Schmidt{-}Thieme, L.
\newblock 2016.
\newblock Hyperparameter optimization machines.
\newblock In {\em Proceedings of the International Conference on Data Science
  and Advanced Analytics (DSAA)},  41--50.
\newblock {IEEE}.

\bibitem[\protect\citeauthoryear{Wolpert}{1992}]{wolpert-nn92a}
Wolpert, D.
\newblock 1992.
\newblock Stacked generalization.
\newblock {\em Neural Networks} 5(2):241--259.

\bibitem[\protect\citeauthoryear{Xu, Hoos, and Leyton-Brown}{2010}]{xu-aaai10a}
Xu, L.; Hoos, H.; and Leyton-Brown, K.
\newblock 2010.
\newblock Hydra: Automatically configuring algorithms for portfolio-based
  selection.
\newblock In Fox, M., and Poole, D., eds., {\em Proceedings of the
  Twenty-fourth National Conference on Artificial Intelligence (AAAI'10)},
  210--216.
\newblock AAAI Press.

\bibitem[\protect\citeauthoryear{Yogatama and
  Mann}{2014}]{yogatama-aistats2014}
Yogatama, D., and Mann, G.
\newblock 2014.
\newblock Efficient transfer learning method for automatic hyperparameter
  tuning.
\newblock In Kaski, S., and Corander, J., eds., {\em Proceedings of the
  Seventeenth International Conference on Artificial Intelligence and
  Statistics (AISTATS)}, volume~33 of {\em {JMLR} Workshop and Conference
  Proceedings},  1077--1085.

\end{thebibliography}

\end{document}